\documentclass[journal]{IAENGtran}
\usepackage{bm}
\usepackage{caption}
\usepackage{subcaption}
\usepackage{amsfonts}
\usepackage{amsmath}
\usepackage{algorithm}
\usepackage{algpseudocode}
\usepackage{float}
\usepackage{tikz}
\usetikzlibrary{shapes,arrows}

\ifCLASSINFOpdf
\DeclareGraphicsExtensions{.pdf,.jpeg,.png}
\else
\usepackage[dvips]{graphicx}
\DeclareGraphicsExtensions{.eps}
\fi

\begin{document}
	%
	\title{K-Means Clustering using Tabu Search \\with Quantized Means}

	\author{Kojo~Sarfo~Gyamfi,
		James~Brusey
		and~Andrew~Hunt
		\thanks{Manuscript received June 27, 2016; revised July 20, 2016. This work was supported
			in part by the National Engineering Laboratory (NEL), United Kingdom.}
		\thanks{Kojo Sarfo Gyamfi is a PhD student with the Faculty
			of Engineering and Computing, Coventry University, Coventry, United Kingdom, (e-mail: gyamfik@uni.coventry.ac.uk).}
		\thanks{James Brusey is a reader with the Faculty
			of Engineering and Computing, Coventry University, Coventry, United Kingdom, (e-mail: aa3172@coventry.ac.uk).}
		\thanks{Andrew Hunt is a professor with the Faculty
			of Engineering and Computing, Coventry University, Coventry, United Kingdom, (e-mail: ab8187@coventry.ac.uk).}
	}

	\maketitle
	
	\pagestyle{empty}
	\thispagestyle{empty}

	\begin{abstract}
		The Tabu Search (TS) metaheuristic has been proposed for K-Means clustering as an alternative to Lloyd's algorithm, which for all its ease of implementation and fast runtime, has the major drawback of being trapped  at local optima. While the TS approach can yield superior performance, it involves a high computational complexity. Moreover, the difficulty in parameter selection in the existing TS approach does not make it any more attractive. This paper presents an alternative, low-complexity formulation of the TS optimization procedure for K-Means clustering. This approach does not require many parameter settings. We initially constrain the centers to points in the dataset. We then aim at evolving these centers using a unique neighborhood structure that makes use of gradient information of the objective function. This results in an efficient exploration of the search space, after which the means are refined. The proposed scheme is implemented in MATLAB and tested on four real-world datasets, and it achieves a significant improvement over the existing TS approach in terms of the intra cluster sum of squares and computational time.
	\end{abstract}
	
	\begin{IAENGkeywords}
		Unsupervised learning, Clustering, K-Means, Tabu Search.
	\end{IAENGkeywords}
	
	\IAENGpeerreviewmaketitle
	\section{Introduction}
	\IAENGPARstart{A}{} common problem in machine learning is the task of having to group a set of $N$ data points or objects into $K$ clusters. This is termed \textit{clustering}. These objects are collected together into a set denoted as $\mathbb{D}$. Clustering can occur in varied settings. As an example, consider the case of classifying $N$ organisms into $K$ different kingdoms based on their features. This can be construed as a clustering problem where the number of features being considered is $d$. In the more general sense, $d$ denotes the dimensionality of the set $\mathbb{D}$. Furthermore, the collection of feature vectors of all the organisms forms the set $\mathbb{D}$, while the clusters denoted as $\mathbb{C}_k$ are represented by the different kingdoms.
	
	In machine learning, clustering falls under the domain of unsupervised learning since there are no class labels to the objects in $\mathbb{D}$. Nonetheless, it can also be performed as a precursor to some supervised learning techniques. An example of this latter application is in the implementation of the radial basis function (RBF) with $K$ centers \cite{1}. 
	
	The clusters, denoted as $\mathbb{C}_k$ $(k=1,...,K)$, are to be determined such that objects in any one cluster are similar to each other, but different from objects in all other clusters. It is assumed that the objects in $\mathbb{D}$ lend themselves to some natural grouping \cite{2}. Otherwise, any partitioning of the data can be considered valid, which would make the problem undefined. However, in the $K$-center RBF, such an assumption is not binding since the objective is to use the $K$ centers as representative points in the dataset for the construction of basis functions. 
	
	Clustering is, however, an ill-posed problem \cite{3} for the following reasons. First, the question of how to tell if any two objects are similar has no definitive answer. To illustrate this, in Fig. 1 (a), the similarity among objects in either of the natural clusters indicated by + or o is based on the distance of a point from the center. On the other hand, in Fig. 1 (b), the closeness of the points to one another provides the measure of similarity among the two natural clusters indicated by + and o. Thus, there is no general similarity measure by which objects are clustered. 
	
	The second reason why the problem of clustering is ill-defined is that the number of clusters $K$ to which the objects must be classified is not known \textit{a priori}. A rough estimate of $K$ is usually assumed to be available from domain expertise or from the distribution of the data. If such an estimate is not available, the common practice is that existing algorithms are run for different $K$. The value of $K$ which minimizes some predefined criterion like the Akaike Information Criterion (AIC) or the Bayes Information Criterion \cite{3} is then chosen. Clustering algorithms may yield poor results if the $K$ chosen is inappropriate \cite{3}.
	
	The most widely used algorithm for clustering in the context of machine learning is Lloyds algorithm, more commonly referred to as K-Means algorithm. It is so called because it essentially computes the $K$ means or centroids of the different clusters. The ease of implementation of the algorithm as well as its fast runtime has accounted for its ubiquity in use. Nevertheless, it has the major drawback of yielding solutions that are only locally optimal, and which may not necessarily be the global optimal solution. For this reason, several other methods have been applied to solving the clustering problem \cite{4}-\cite{7}. Notable among these is the approach of Al-Sultan \cite{7} which is based on the Tabu Search (TS) algorithm developed by Glover \cite{8}. We henceforth refer to this approach, i.e. \cite{7} (our reference work) as the Tabu Search Clustering (TSC) algorithm. The performance reported was shown to be superior to that of the K-Means algorithm. 
	
	The TS algorithm is a metaheuristic procedure that accepts an initial solution as input, and performs a local search using neighborhood and memory structures until some stopping criterion is met. It is able to escape local minima by allowing for solutions that do not improve the objective function. TS has been applied in solving varied problems including the traveling salesman problem (TSP) \cite{9} and signal detection in multiple input multiple output (MIMO) antenna systems \cite{10}. However, with regards to the clustering problem, the high computational complexity and difficulty in parameter selection required in the TS approach does not make it an attractive alternative to the K-Means algorithm.
	
	Our main contributions in this paper are as follows: 
	\begin{enumerate}		
		\item We introduce a quantized means TS scheme for solving the clustering problem. We target the optimization of the $K$ centers by evolving them through a series of neighboring solutions in such a manner that leads to an efficient exploration of the search space. This procedure is well described in Section IV. The scheme requires only two parameters to be set, and is of a relatively low complexity.
		
		\item We present experimental results obtained from the proposed approach on some test datasets (Section V).
	\end{enumerate}
	
	\section{Problem Description}
	For the purpose of this paper, we assume that the number of clusters $K$ is given. We refer the reader to the works by Hamerly et al. \cite{11} and Pan et al. \cite{12} for a detailed treatment on how to choose $K$. The dataset $\mathbb{D}$ is assumed to come from a mixture distribution where the mixture component label (which is the cluster index) for any object in the dataset is hidden. In the most general sense, an object can belong to more than one cluster. Thus, for such an interpretation, the problem of clustering is simply that of finding the clusters to which an object belongs with a high probability. Mathematically, this can be stated concisely as maximizing the following probability for different models $\textbf{S}$ for a given $\textbf{x}_n \in \mathbb{D}$:
	\begin{equation}
		p(\textbf{x}_n \mid \textbf{S})=\sum_{k=1}^{K}w_kp(\textbf{x}_n \mid \mathbb{C}_k,\textbf{S})
	\end{equation}
	where $\textbf{S}$ comprises the cluster memberships for all the objects in the dataset, as well as the mixture weights $w_k$.
	
	Maximizing (1) requires knowledge of the cluster memberships and the mixture weights, as well as knowledge of the mixture distribution. In general, none of these is known, and so the following set of simplifying assumptions \cite{4} are made in practice.
	\begin{enumerate} 
		\item Each object in $\mathbb{D}$ belongs to a single cluster;
		\item Each cluster is distributed as a multivariate Gaussian;
		\item The mixture components have equal weights $w_k$.
	\end{enumerate}
	
	\begin{figure}[tbph]
		\centering
		\begin{subfigure}{0.5\textwidth}
			\centering
			\includegraphics[width=\textwidth, height=60mm]{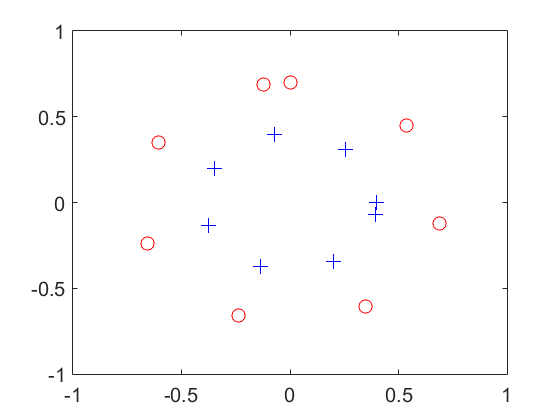}
			\caption{}
			\label{m1}
		\end{subfigure}
		\begin{subfigure}{0.5\textwidth}
			\centering
			\includegraphics[width=\textwidth, height=60mm]{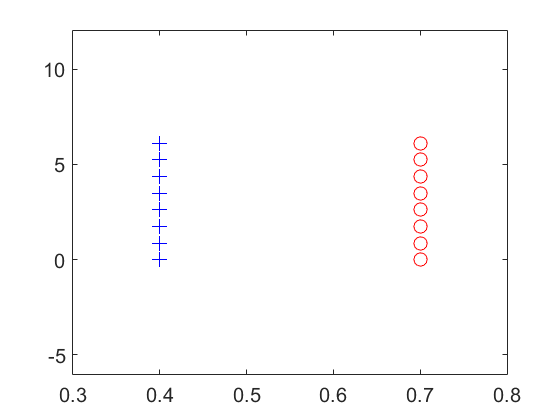}
			\caption{}
			\label{m2}
		\end{subfigure}
		\caption{Similarity Measures}
	\end{figure}
	
	The consequence of the above assumptions is that the similarity measure is now based on the Euclidean norm so that points closest to each other in Euclidean space are grouped under one and only one cluster. It is conceivable that a dataset may have some similarity measure other than the Euclidean distance. Indeed, \cite{13}-\cite{15} explore the use of other distance measures for clustering. Yet, for some datasets, an appropriate representation of the points can make the Euclidean distance measure valid. As an example, transformation of the points in Fig. 1 (a) into polar co-ordinates yields the representation in Fig. 1 (b) which has the Euclidean distance as a valid similarity measure. The clustering problem then yields itself to a treatment as a mathematical optimization whose aim is to minimize a parameter $J$ known as the intra cluster sum of squares (ICSS) or the distortion \cite{4}. This may be stated as:
	
	\begin{equation}
		\min J=\sum_{k=1}^{K}\sum_{\textbf{x}_n\in \mathbb{C}_k}{\|\textbf{x}_n-\bm{\mu}_k\|}^2
	\end{equation}
	where $\textbf{x}_n \in \mathbb{C}_k$ are all data points in cluster $k$ and $\bm{\mu}_k$ is the mean or center of that $k$th cluster. This is the problem termed as K-Means clustering. It must be mentioned that neither the cluster memberships nor the means are known. Thus, this problem is computationally difficult, and is NP-hard \cite{6}. 
	
	The K-Means algorithm provides an efficient way of solving (2). It is based on the observation that the optimal placement of the $K$ centers is at the centroids of the respective clusters. The algorithm is typically initialized with some random means, usually chosen from objects in the dataset $\mathbb{D}$. Since the K-Means algorithm is a special case of the Expectation-Maximization (EM) algorithm \cite{4}, it proceeds in two stages namely, the expectation and the maximization stages.
	\begin{enumerate}
		\item Expectation: Compute the centroid of each cluster:
		\begin{equation}
			\bm{\mu}_k=\frac{1}{N_k}\sum_{\textbf{x}_n\in \mathbb{C}_k}\textbf{x}_n
		\end{equation}
		where $N_k$ is the number of objects in the $k$th cluster.
		\item Maximization: Compute the cluster memberships:
		\begin{align}
			\mathbb{C}_k = \left\{ \textbf{x}_n: \|\textbf{x}_n-\bm{\mu}_k\|^2 < \|\textbf{x}_n-\bm{\mu}_l\|^2, \right. \nonumber \\ 
			\quad \left. l=1,...,K, \quad l\not=k \right\}
		\end{align}
	\end{enumerate}
	The expectation-maximization steps are carried out iteratively until there is no cluster change, at which point the algorithm is terminated.
	
	The major drawback of the K-Means algorithm is as follows. First, the objective function of (2) is non-convex and may thus have several local minima. Therefore, being only a local search method, the K-Means algorithm is not guaranteed to find a global minimum; it often yields solutions that are only locally optimal. This is due in part to the nature of the stopping criterion. The algorithm terminates when there is no change in cluster memberships; this period corresponds to a local minimum. It makes no provision to consider other local minima which may be present in other areas of the search space. Again, as with all local search methods, the performance of the K-Means algorithm is directly tied to the quality of the initial solution. If this solution is poor, i.e., if it is too far away from the global optimum, the algorithm may likely converge to a local minimum.
	
	Our approach is motivated by the above limitation, and is based on the TS algorithm in \cite{8}.
		
	\section{Tabu Search}
	Tabu Search is a metaheuristic technique used for combinatorial optimization. It does not require the optimization problem to be convex. The algorithm makes use of neighborhood structures to explore the search space. It also utilizes a short term memory structure called a \textit{tabu}, which is essentially a list of forbidden moves or solutions. Tabus prevent the back and forth movements between solutions that have already been considered in the search, a phenomenon called \textit{cycling}. Moreover, TS allows for moves to solutions that do not yield any improvement in the objective function. It does so with the view that the poor solution may lead to a better one at a later time in the search. Thus, it is able to escape from local minima. TS keeps in memory the best solution found at any point in the search, and returns that solution when the algorithm is terminated. In its most basic form, it follows the procedure outlined below:
	
	\begin{enumerate}
		\item Select an initial solution $\textbf{M}^{(0)}$. This solution can be randomly generated or obtained by more formal means. Set $\textbf{M}_c$ and $\textbf{M}_b$ to $\textbf{M}^{(0)}$. $\textbf{M}_c$ and $\textbf{M}_b$ are the current and best solutions respectively.
		
		\item Evaluate the objective function $J$ for the current solution $\textbf{M}_c$.
		
		\item Find neighboring solutions of $\textbf{M}_c$. Let $\textbf{V}$ denote this set. The neighbors of $\textbf{M}_c$ are all those solutions that are similar to, but differ in a minor aspect from $\textbf{M}_c$.
		
		\item Find the set of solutions in $\textbf{V}$ that are not in the Tabu list $\textbf{T}$. Let this set be denoted by $\textbf{V}\setminus\textbf{T}$. The Tabu is a list of solutions or moves that have already been considered in the search. Tabus, as algorithmic structures, force the algorithm to other areas of the search space, thus enhancing the diversification of the search.
		
		\item Evaluate the objective function for all the solutions in $\textbf{V}\setminus\textbf{T}$. Find the best solution among this set. Let this be $\textbf{M}_n$.
		
		\item If $J_n < J_b$, let $\textbf{M}_b=\textbf{M}_n$. $J_n$ and $J_b$ are the objective function evaluations of $\textbf{M}_n$ and $\textbf{M}_b$ respectively.
		
		\item Put the solution $\textbf{M}_c$ into the Tabu list, and let $\textbf{M}_n$ be the new current solution $\textbf{M}_c$. If the maximum number of iterations (which is chosen beforehand) has elapsed, terminate. Else, go to Step 3.
	\end{enumerate}

	\section{Quantized Means TS Clustering}
	In this section, we discuss the proposed algorithm. As with any TS implementation, the Quantized Means TS Clustering follows the skeleton of the description of the TS algorithm in Section III with the following modifications and specificities.
	
	\subsection{Search Space}
		
	In this formulation, a vector $\textbf{M}^{(0)}$ defined as $\textbf{M}^{(0)}=[\bm{\mu}_1^T,...,\bm{\mu}_K^T]^T$ is considered as the initial solution, where $\bm{\mu}_1,...,\bm{\mu}_K$ are $K$ randomly chosen observations from the dataset $\mathbb{D}$. $\textbf{M}^{(0)}$ is then assigned to $\textbf{M}_c$. 
	
	To navigate the search space then, neighbors of $\textbf{M}_c$ have to be found. Neighboring solutions are typically drawn from a finite set that includes the current solution itself. Alternatively, they can be obtained via a simple transformation of the current solution. It is worth mentioning that in the context of TS, neighboring solutions are not necessarily those that are closest to the current solution. 
	
	To obtain neighbors of $\textbf{M}_c$, we change its individual components i.e. $\bm{\mu}_k$ $(k=1,...,K)$, by replacing them with some new means or centers. However, the means are real-valued in general, and do not constitute any finite set. Therefore, the set of all possible neighbors obtained in this manner is necessarily an infinite set. This set is the feasible search space. The fact of the search space being infinite makes TS ill-suited to optimizing $\textbf{M}_c$, since TS is used for combinatorial optimization.
	
	A finite subset of the search space is thus necessary. For this reason, the proposed scheme makes the assumption that the $K$ means take on values exclusively from objects in the dataset $\mathbb{D}$. We refer to this as \textit{quantized means}. Our proposed algorithm is divided into two stages, namely, exploration and refinement; we make the aforementioned assumption on the means only in the initial exploration stage. Thus, in the exploration stage, for $k=1,...,K$, $\bm{\mu}_k \in \textbf{M}_c$ is replaced with some other point $\textbf{x}$ taken from the dataset. This procedure yields the neighboring solution denoted as $\textbf{M}_n$. More specifically, for every $k \in \left\lbrace 1,...,K\right\rbrace $, we constrain the point $\textbf{x}$ to the $k$th cluster $\mathbb{C}_k$ (which is a subset of the dataset). This quantization of the means makes the problem formulation combinatorial. Nevertheless, the resulting set of all possible combinations of $\textbf{M}_c$ (i.e. the search space denoted as $\mathbb{W}$), although finite, is still large.
	
	\tikzstyle{decision} = [diamond, draw, fill=blue!20, 
	text width=5em, text badly centered, node distance=3cm, inner sep=0pt]
	\tikzstyle{block} = [rectangle, draw, fill=blue!20, 
	text width=8em, text centered, rounded corners, minimum height=4em]
	\tikzstyle{line} = [draw, -latex']
	\tikzstyle{cloud} = [draw, ellipse,fill=red!20, node distance=3cm,
	minimum height=2em]

	\subsection{Neighborhood Construction}	
	Due to the large size of $\mathbb{W}$, one has to choose only $R$ ($R \ll \left| \mathbb{W}\right| $) points from the set $\mathbb{W}$ via a simple transformation of the solution $\textbf{M}_c$ and consider those as the neighbors of $\textbf{M}_c$ in any one iteration of the TS algorithm. The difficulty, however, is in the choice of which $R$ neighbors. 
	
	If we randomly select neighbours, the search is unguided and thus likely to be slow to converge on the optimal solution. A simple guiding mechanism might be to choose nearest neighbours. However, this seems a poor choice intuitively because for many cases it will cause no change to the clustering and where it does, it might not be a change in the right direction.
	
	\subsubsection*{Analytic Neighbors}
	We therefore use the gradient information of the objective function to guide the neighbor selection. In any TS iteration, we choose $R$ points in $\mathbb{W}$ that result in the steepest descent along the trajectory of the objective function. We consider the selection of a single neighbor, i.e $R=1$ in this paper. The following approach is then taken to find one high-quality neighbor of $\textbf{M}_c$. Since the objective function of (2) is non-convex, the aim is to find a neighbor $\textbf{M}_n\in\mathbb{W}$ that corresponds to a local minimum of $J$. A necessary and sufficient condition for this is to have the gradient of $J$ to be zero at the local minimum, i.e.
	
	\begin{equation}
		\nabla_{\textbf{M}_c}J=\textbf{0}
	\end{equation}
	where the notation $\nabla$ represents the gradient. By definition,
	\begin{equation}
		\nabla_{\textbf{M}_c}J= \frac{\partial J}{\partial \bm{\mu}_1}\textbf{e}_1 +... + \frac{\partial J}{\partial \bm{\mu}_K}\textbf{e}_K=\textbf{0}
	\end{equation}
	where $\textbf{e}_k$ $(k=1,...,K)$ is a unit vector in the $\bm{\mu}_k$ direction. By (6), it has implicitly been assumed that each $\bm{\mu}_k$ is independent of the other. This assumption is not generally true of the K-Means algorithm, as a change in some $\bm{\mu}_k$ may change the cluster memberships and hence change the location of the other means. However, in the exploration stage of our proposed algorithm, the means are not defined as the cluster centroids as in (3), but are chosen independently of each other in the procedure below. This permits the evaluation of the $K$ partial derivatives independently as:
	\begin{equation}
		\frac{\partial J}{\partial \bm{\mu}_k}=0, \quad \forall k=1,...,K
	\end{equation}
	However, since $\bm{\mu}_k$ has been constrained to the dataset $\mathbb{D}$, the change in the means $\partial \bm{\mu}_k$ being considered is not necessarily infinitesimal. For this reason, we approximate the partial derivative in (6) as a partial difference quotient as:
	\begin{equation}
		\frac{\partial J}{\partial \bm{\mu}_k}\approx\frac{\Delta J}{\Delta \bm{\mu}_k}\approx 0, \quad \forall k=1,...,K
	\end{equation}
	which can be evaluated from first principles as follows:
	\begin{equation}
		J=\sum_{i=1}^{K}\sum_{\textbf{x}_n\in \textbf{C}_i}{\|\textbf{x}_n-\bm{\mu}_i\|}^2
	\end{equation}
	A change in the mean $\Delta\bm{\mu}_i$ would then cause a change $\Delta J$ in the objective function, i.e.,
	\begin{equation}
		J+\Delta J=\sum_{i=1}^{K}\sum_{\textbf{x}_n\in \textbf{C}_i}{\|\textbf{x}_n-(\bm{\mu}_i+\Delta\bm{\mu}_i)\|}^2
	\end{equation}
	\begin{align}
		& J+\Delta J =\nonumber \\
		& \sum_{i=1}^{K}\sum_{\textbf{x}_n\in \textbf{C}_i} \big ({\|\textbf{x}_n-\bm{\mu}_i\|}^2-2{(\textbf{x}_n-\bm{\mu}_i)}^T\Delta\bm{\mu}_i+{\|\Delta\bm{\mu}_i\|}^2 \big )
	\end{align}
	\begin{equation}
		\Delta J=\sum_{i=1}^{K}\sum_{\textbf{x}_n\in \textbf{C}_i}\big (-2{(\textbf{x}_n-\bm{\mu}_i)}^T\Delta\bm{\mu}_i+{\|\Delta\bm{\mu}_i\|}^2 \big )
	\end{equation}

	For the purpose of evaluating (8), $\Delta\bm{\mu}_i=0$ for $i\not=k$ due to the assumption of the independence of the means, hence
	\begin{equation}
		\Delta J=\sum_{\textbf{x}_n\in \mathbb{C}_k}\big (-2{(\textbf{x}_n-\bm{\mu}_k)}^T\Delta\bm{\mu}_k+{\|\Delta\bm{\mu}_k\|}^2 \big )
	\end{equation}
	\begin{equation}
		\frac{\Delta J}{\Delta\bm{\mu}_k}=\sum_{\textbf{x}_n\in \mathbb{C}_k}\big(-2{(\textbf{x}_n-\bm{\mu}_k)}+\Delta\bm{\mu}_k \big )\approx 0
	\end{equation}
	where $\Delta\bm{\mu}_k=\textbf{x}-\bm{\mu}_k$.
		
	In order to evaluate (14), we find $\textbf{x} \in \mathbb{C}_k$ that minimizes (13), having constrained the means to the dataset $\mathbb{D}$. This minimizing parameter is denoted as $\textbf{x}_k^*$. The neighboring solution $\textbf{M}_n$ is then the aggregation of all $\textbf{x}_k^*$ $(k=1,...,K)$, i.e. $\textbf{M}_n=[\textbf{x}_1^{*T},...,\textbf{x}_K^{*T}]^T$. It must be noted that if the means were to be unconstrained to $\mathbb{D}$, (7) could be evaluated directly instead of solving (14), and the solution of (7) would be the centroids of the clusters, which is essentially what the K-Means algorithm evaluates. However, the initial assumption on the means would be violated, and there would still be the risk of getting trapped at local minima. Rather, this procedure allows for the consideration of solutions that worsen the objective function $J$ since (13) would not always yield negative values, thereby escaping from local minima. 
	
	The intuitive alternative of finding the cluster centroids, and then quantizing them to the dataset $\mathbb{D}$ introduces a quantization loss and yields an inferior performance to the procedure described.
	
	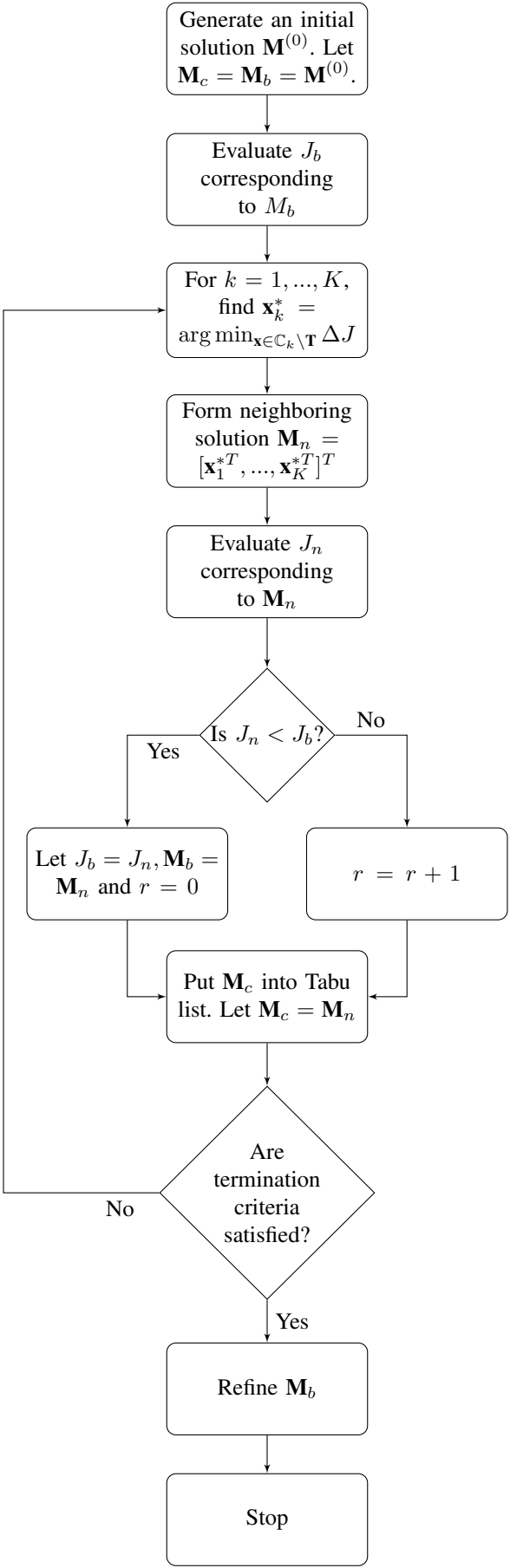
\begin{figure}[tbph]
		\begin{tikzpicture}[node distance = 2cm, auto]
		\node [block, fill=white] (init) {Generate an initial solution $\textbf{M}^{(0)}$. Let $\textbf{M}_c=\textbf{M}_b=\textbf{M}^{(0)}$.};
		\node [block, below of=init, fill=white] (Jb) {Evaluate $J_b$ corresponding to $M_b$};
		\node [block, below of=Jb, fill=white] (identify) {For $k=1,...,K$, find $\textbf{x}_k^*= \arg \min_{\textbf{x} \in \mathbb{C}_k\setminus\textbf{T}} \Delta J$};
		\node [block, below of=identify, fill=white] (evaluate) {Form neighboring solution  $\textbf{M}_n=[\textbf{x}_1^{*T},...,\textbf{x}_K^{*T}]^T$ };
		\node [block, below of=evaluate, fill=white] (Jn) {Evaluate $J_n$ corresponding to $\textbf{M}_n$};
		\node [decision, below of=Jn, node distance=2.5cm, fill=white] (decide1) {Is $J_n<J_b$?};
		\node [block, below left of=decide1, node distance =3cm, fill=white] (expert1) {Let $J_b=J_n, \textbf{M}_b=\textbf{M}_n$ and $r=0$};
		\node [block, below right of=decide1, node distance =3cm, fill=white] (expert2) {$r=r+1$};
		\node [block, below of=decide1, node distance =4cm, fill=white] (current) {Put  $\textbf{M}_c$ into Tabu list. Let $\textbf{M}_c=\textbf{M}_n$};
		\node [decision, below of=current, fill=white] (decide3) {Are termination criteria satisfied?};
		\node [block, below of=decide3, node distance =3cm, fill=white] (refine) {Refine $\textbf{M}_b$};
		\node [block, below of=decide1, node distance=12cm, fill=white] (stop) {Stop};
		\path [line] (init) -- (Jb);
		\path [line] (Jb) -- (identify);
		\path [line] (identify) -- (evaluate);
		\path [line] (evaluate) -- (Jn);
		\path [line] (Jn) -- (decide1);
		\path [line] (decide1) -| node  [near start] {Yes} (expert1);
		\path [line] (decide1) -| node  [near start] {No} (expert2);
		\path [line] (expert1) |- (current);
		\path [line] (expert2) |- (current);
		\path [line] (current) -- (decide3);
		\path [line] (decide3) -- node {Yes} (refine);
		\path[line] (decide3) -- node  [near start] {No} ++(-4cm,0) |- (identify);
		\path [line] (refine) -- (stop);
		
		\end{tikzpicture}
		\caption{Proposed Algorithm}
	\end{figure}
	
	Once the means have been computed, the cluster memberships can then be determined from:
	\begin{align}
	\mathbb{C}_k = \left\{ \textbf{x}_n: \|\textbf{x}_n-\textbf{x}_k^*\|^2 < \|\textbf{x}_n-\textbf{x}_l^*\|^2, \right. \nonumber \\ 
	\quad \left. l=1,...,K, \quad l\not=k \right\}
	\end{align}
	from which the objective function in (1) can be evaluated.

	\subsection{Tabu}
	The Tabu structure used in this formulation is a list of all the means $\textbf{x}_k^*$ that have been considered in the search. Since there are $K$ means, the Tabu considered is an array with $K$ rows whose column length increases as the TS algorithm proceeds. If for some $k$, $\textbf{x}_k^*$ obtained from minimizing (13) is in the Tabu list, it is discarded and the next point $\textbf{x} \in \mathbb{C}_k$ in increasing order of their $\Delta J$ evaluation is chosen. If all points in the $k$th cluster are in the Tabu list for any $k$, the last entry in the $k$th row of the Tabu list is deleted in order to allow for at least one solution to be valid.

	\subsection{Termination Criterion}
	The termination criterion employed in the proposed algorithm is two-fold. First, after the maximum number of TS iterations $IT_{max}$ has been reached, the algorithm is terminated. Secondly, there is an early termination criterion whereby the algorithm is cut off after a predefined number of iterations (called the cut-out parameter $r_{max}$) within which there is no improvement in the best found solution $\textbf{M}_b$. This is an indicator of the convergence of the algorithm. The early termination is done on the assumption that the global minimum may have already been achieved. In order for this assumption to be mostly valid, the neighboring solutions generated in any iteration must not be random. Otherwise, there is a good chance the global optimal solution would be found in any TS iteration. Therefore, the process of generating $R$ random neighbors of $\textbf{M}_c$ from $\mathbb{W}$ would not yield good results with regards to the early termination. The early termination cuts down the computational complexity as the algorithm does not need to be run for all $IT_{max}$ iterations.

	\subsection{Refinement}
	The essence of the initial restriction on the means $\bm{\mu}_k$ to belong to the finite set $\mathbb{D}$ is to make the optimization problem combinatorial, and enable the efficient exploration of the search space. Once that has been achieved at the end of the TS algorithm, the means can then be unconstrained. As a result, the components of the best found solution $\textbf{M}_b$ are recomputed as the centroids of the clusters obtained at the end of the TS algorithm. Alternatively, one may use $\textbf{M}_b$ as an initial solution to the K-Means algorithm to obtain a refined solution.
	
	The proposed TS scheme (i.e. using the analytic neighborhoods) is illustrated in the flow chart of Fig. 2.

\section{Simulations and Results}
For our simulations, we use four real-world datasets namely: the Bavaria Postal code dataset \cite{16} (for two different values of $K$), the Fisher's \textit{Iris} dataset, the Glass Identification dataset, and the normalized Cloud dataset \cite{17} (also for two different values of $K$). These datasets are chosen to cut across a wide range of $d$, $K$, and $N$ values. We simulate our proposed scheme in MATLAB on an Intel Core i5-2400 processor using the following parameters: $IT_{max}=400$ and $r_{max}=0.25IT_{max}$. We compare the performance of this scheme to the TSC, the K-Means++ \cite{18}, and the K-Means algorithms in terms of the objective function of (2) and the time taken for completion. We use the following parameter settings for the TSC algorithm: $NTS=20,\mathit{MTLS}=15,IT_{max}=1000, P=0.95$ as suggested by Al-Sultan \cite{5}. For each dataset, we run each algorithm $100$ times, and provide the worst, average and best objective function values, as well as the average time for completion. The results of our simulations are summarized in Tables I-VI.

\begin{table}[H]
	\centering
	\caption{Iris Flower dataset\\
		$N=150, K=3, d=4$}
	{\begin{tabular}{|c|c|c|c|c|}
			\hline
			Algorithm & Worst $J$ & Average $J$ & Best $J$ & Time (s)\\
			\hline
			K-Means & $ 145.76 $ & $ 90.50 $ & $ 78.85 $ & $ 0.17 $ \\
			K-Means++ & $ 145.45 $ & $ 80.16 $ & $ 78.85 $ & $ 0.18 $ \\
			TSC & $ 310.48 $ & $ 282.94 $ & $ 249.93 $ & $ 72.98 $ \\
			Proposed & $ 78.86 $ & $ \textbf{78.85} $ & $ 78.85 $ & $ 3.58 $ \\
			\hline
	\end{tabular}}{}		
	\label{iris}
\end{table}

\begin{table}[H]
	\centering
	\caption{Glass dataset\\
		$N=214, K=6, d=9$}
	{\begin{tabular}{|c|c|c|c|c|}
			\hline
			Algorithm & Worst $J$ & Average $J$ & Best $J$ & Time (s)\\
			\hline
			K-Means & $ 580.02 $ & $ 394.32 $ & $ 336.29 $ & $ 0.36 $ \\
			K-Means++ & $ 480.82 $ & $ 376.25 $ & $ 336.06 $ & $ 0.38 $ \\
			TSC & $ 928.51 $ & $ 904.08 $ & $ 873.68 $ & $ 68.26 $ \\
			Proposed  & $ 382.13 $ & $ \textbf{352.28} $ & $ 338.75 $ & $ 5.33 $ \\
			\hline
	\end{tabular}}{}		
	\label{glass}
\end{table}
		
\begin{table}[H]
	\centering
	\caption{Bavaria dataset\\
		$N=89, K=4, d=3$}
	{\begin{tabular}{|c|c|c|c|c|}
			\hline
			Algorithm & Worst $J$ & Average $J$ & Best $J$ & Time (s)\\
			\hline
			K-Means & $ 2.79$e$+11 $ & $ 2.67$e$+11$ & $ 1.04$e$+11$ & $ 0.25 $ \\
			K-Means++ & $ 2.79$e$+11$ & $ 1.16$e$+11$ & $1.04$e$+11$ & $ 0.15 $ \\
			TSC & $ 4.38$e$+11$ & $ 4.10$e$+11$ & $3.84$e$+11$ & $55.66$ \\
			Proposed & $ 1.05$e$+11$ & $ \textbf{1.05$e$+11} $ & $ 1.04$e$+11$ & $ 3.84 $ \\
			\hline
	\end{tabular}}{}	
	\label{bav1}
\end{table}
			
\begin{table}[H]
	\centering
	\caption{Bavaria dataset\\
		$N=89, K=5, d=4$}
	{\begin{tabular}{|c|c|c|c|c|}
			\hline
			Algorithm & Worst $J$ & Average $J$ & Best $J$ & Time (s)\\
			\hline
			K-Means & $ 2.62$e$+11$ & $ 2.56$e$+11$ & $ 0.74$e$+11$ & $ 0.35 $ \\
			K-Means++ & $ 8.65$e$+10$ & $ \textbf{6.83$e$+10} $ & $ 5.98$e$+10$& $ 0.23 $ \\
			TSC & $3.76$e$+11$ & $ 3.17$e$+11$ &$ 2.16$e$+11$ & $ 64.19 $ \\
			Proposed & $8.07$e$+10$ & $8.02$e$+10$ & $ 5.98$e$+10$ & $ 6.57 $ \\
			\hline
	\end{tabular}}{}		
	\label{bav2}
\end{table}

\begin{table}[H]
	\centering
	\caption{Cloud dataset\\
		$N=1024, K=10, d=10$}
		{\begin{tabular}{|c|c|c|c|c|}
				\hline
				Algorithm & Worst $J$ & Average $J$ & Best $J$ & Time (s)\\
				\hline
				K-Means & $ 1646.53$ & $ 1580.19$ & $ 1504.08$ & $ 6.57 $ \\
				K-Means++ & $ 1664.17$ & $ 1550.40$ & $ 1504.58$ & $ 7.21 $ \\
				TSC & $ 9140.61$ & $ 9058.90$& $ 8895.18$& $ 810.97 $ \\
				Proposed & $ 1596.17$ & $ \textbf{1530.95} $ & $ 1503.18$ & $ 29.32 $ \\
				\hline
			\end{tabular}}{}		
	\label{cloud1}
\end{table}
					
\begin{table}[H]
	\centering
	\caption{Cloud dataset\\
		$N=1024, K=25, d=10$}
		{\begin{tabular}{|c|c|c|c|c|}
				\hline
				Algorithm & Worst $J$ & Average $J$ & Best $J$ & Time (s)\\
				\hline
				K-Means & $ 1105.65$ & $ 944.94$ & $ 847.10$ & $ 15.40 $ \\
				K-Means++ & $ 937.12$& $ 848.73$ & $ 820.46$ & $ 19.28$ \\
				TSC & $ 8897.38$ & $ 8746.27$& $ 8576.96$ & $ 814.79 $ \\
				Proposed & $ 920.42$ & $ \textbf{845.25} $ & $ 816.87$ & $ 83.76 $ \\
				\hline
			\end{tabular}}{}		
	\label{cloud2}
\end{table}

From the tables, it can be seen that the proposed scheme achieves the best average objective function in five of the six tests performed. The best average objective function for all the tests have been highlighted in boldface. The proposed scheme consistently outperforms the TSC algorithm in terms of the computational time, average, best and worst objective function values. Specifically, on the Cloud dataset for $K=25$, our algorithm achieves as much as $90\%$ improvement on the average $J$, while doing so $90\%$ faster. It must be noted that both the proposed scheme and the TSC algorithm can actually be used to obtain lower values of $J$ than the ones reported, by increasing the value of $IT_{max}$ (and $NTS$ in the case of the TSC). However, that would be at the expense of greater computational time. 

Compared to the K-Means and K-Means++, our algorithm also performs favorably. In particular, it outperforms the K-Means algorithm by as much as $69\%$ in terms of the average $J$ on the Bavaria postal code dataset for $K=5$. Compared to the K-Means++ algorithm, our approach achieves a marginal performance improvement in the average $J$, reaching to $9\%$ on the Bavaria postal code dataset for $K=4$. The proposed scheme also achieves the lowest worst objective function as compared to the K-Means and K-Means++ algorithms on all datasets. However, in terms of the rate of convergence, the K-Means++ algorithm is shown to be the best.

\section{Related Work}
The TS algorithm has been applied to the K-Means clustering problem with a different formulation by Al-Sultan \cite{7} where a candidate solution in the form of an array of length $N$ is used. This array denoted as $\textbf{A}_c$ is made up of the cluster indices of all the $N$ objects in $\mathbb{D}$. In order to obtain neighboring solutions (also known as trial solutions), the cluster indices in $\textbf{A}_c$ are changed according to some criterion. This method can lead to bad cluster memberships. This is because while two close objects in the dataset may show a tendency of belonging to one cluster, this scheme may assign different cluster indices to them. The algorithm also involves setting the following parameters: the number of trial solutions $NTS$, the maximum tabu list size $\mathit{MTLS}$, the maximum number of TS iterations $IT_{max}$, and a probability threshold $P$. Extensive parametric study has to be carried out for a particular dataset in order to obtain the optimal values.

The TS clustering algorithm in \cite{19} discusses essentially the same procedure as the TSC algorithm with two additional neighborhood structures presented. The process of generating neighboring solutions in both of these algorithms is largely random, causing the algorithm to behave to some degree like a random search with memory. The implication of this randomness is that the global optimal solution has an equal chance of being generated in the first TS iteration as it has in the $IT_{max}$th iteration. Consequently, the probability that a global solution may have been found after $r_{max}$ iterations of non-improving solutions is rather low. Thus, the early termination described in Section IV-B cannot be applied to these algorithms without a significant performance loss.

The TS algorithm has also been applied to the Fuzzy C-Means clustering problem \cite{20}, where an object in the dataset $\mathbb{D}$ can belong to more than one cluster to varying degrees. The TS procedure taken in that formulation aims at optimizing the cluster means, which is similar to the approach taken in our proposed scheme. However, that is as far as the similarity goes. While our scheme constrains the means to objects in the dataset and use gradient information to generate a new neighbor, this approach generates neighboring means by perturbing the current mean along a random direction.

Other TS approaches for clustering includes the \textit{packing-releasing} algorithm \cite{21} which is also based on \cite{7}, but with the following fundamental difference: a pair of objects in the dataset that are close to each other are packed together and treated as one object. These packed objects are later released. This procedure reduces the size of the search space and guides the search to a local minimum more quickly.

While we have assumed in this work that the number of clusters $K$ is known beforehand, the evolution-based tabu search algorithm \cite{12} uses TS for the determination of the number of clusters in the dataset, by considering $K$ as another variable to be optimized in the TS procedure.

\section{Conclusion}
In this paper, we have presented an efficient Tabu Search procedure for solving the K-Means clustering problem. This involves constraining the $K$ means to objects in the dataset, and optimizing these means via a series of neighbors that are obtained using gradient information of the objective. We have compared the proposed scheme to an existing TS algorithm as well as the K-Means and K-Means++ algorithms. We have shown that this approach performs favorably with these well-known algorithms, as well as not requiring too many parameter settings. This is a promising result for a lot of machine learning applications that use K-Means clustering. We note, however, that the nature of the tabu structure used in our TS implementation might require a large memory, especially for big datasets where the maximum number of TS iterations is correspondingly large. For this reason, ongoing work is focused on identifying a more compact representation of the entries in the tabu structure and consequently reducing the runtime of the algorithm.

\ifCLASSOPTIONcaptionsoff
  \newpage
\fi



%

\end{document}